\documentclass{article}
\usepackage[final,nonatbib]{nips_2016} 

\usepackage[utf8]{inputenc} 
\usepackage[T1]{fontenc}    
\usepackage{hyperref}       
\usepackage{url}            
\usepackage{booktabs}       
\usepackage{amsfonts}       
\usepackage{nicefrac}       
\usepackage{microtype}      
\usepackage{blindtext, graphicx}
\usepackage{amsmath,amssymb}
\usepackage{bbm}
\usepackage[backend=bibtex,style=nature]{biblatex}
\usepackage{subfig}
\usepackage{color}
\usepackage{multicol}
\usepackage{amsmath}

\DeclareMathOperator{\erfc}{erfc}

\begin{document}

\title{Learning to Communicate: Channel Auto-encoders, Domain Specific Regularizers, and Attention}

\author{
  Timothy J. O'Shea \\
  Virginia Tech ECE\\
  Arlington, VA \\
  \texttt{oshea@vt.edu} 
  \And
  Kiran Karra \\
  Virginia Tech ECE\\
  Arlington, VA \\
  \texttt{kiran.karra@vt.edu} 
  \AND
  T. Charles Clancy \\
  Virginia Tech ECE\\
  Arlington, VA \\
  \texttt{tcc@vt.edu} 
}

\maketitle

\begin{abstract}
We address the problem of learning efficient and adaptive ways to communicate binary information over an impaired channel.  We treat the problem as reconstruction optimization through impairment layers in a channel autoencoder and introduce several new domain-specific regularizing layers to emulate common channel impairments.  We also apply a radio transformer network based attention model on the input of the decoder to help recover canonical signal representations.  We demonstrate some promising initial capacity results from this architecture and address several remaining challenges before such a system could become practical.
\end{abstract}

\section{Introduction}
Radio communication theory has long sought to find algorithms to attain efficient transfer of information over a variety of communications channels.   These range from thermal noise limited channels exhibiting relatively simple Gaussian noise like behaviors to much more complex channels exhibiting multi-path fading, impulse noise, spurious or continuous jamming, and numerous other complex impairments.  Information theory \cite{Shannon} and analysis of modulation schemes gives us bounds on achievable Information density and bit error rates for given transmission rates, bandwidths and signal to noise ratios (SNRs), but does not tell us how to achieve them.

Throughout the years we have achieved numerous discrete operation steps along these bounds allowing efficient operation at specific SNR values.  Rate matching and code adaptation has allowed us to continue to operate very close to this capacity curve by choosing many discrete operating points, but many of them are computationally complex in practice and require expensive hardware or DSP software to leverage in mobile radio systems. 

By taking the approach of unsupervised learning of an end-to-end communications system by optimization of reconstruction cost in a channel auto-encoder with a set of domain appropriate representative channel regularizers, we seek to learn new methods of modulation which blur the lines between modulation and error correction, providing similar SNR to bit error rate performance (BER) while achieving lower computational complexity requirements at runtime.  We believe this will lead to a new class of communications systems with a greater generalization and ability to adapt to impairments for which traditional communication systems can not easily.  The potential impact of such a system on the development, deployment and capabilities of wireless systems holds enormous potential.

\section{Channel Auto-encoders}

\begin{figure}[ht!]
  \centering
      \includegraphics[width=0.6\textwidth]{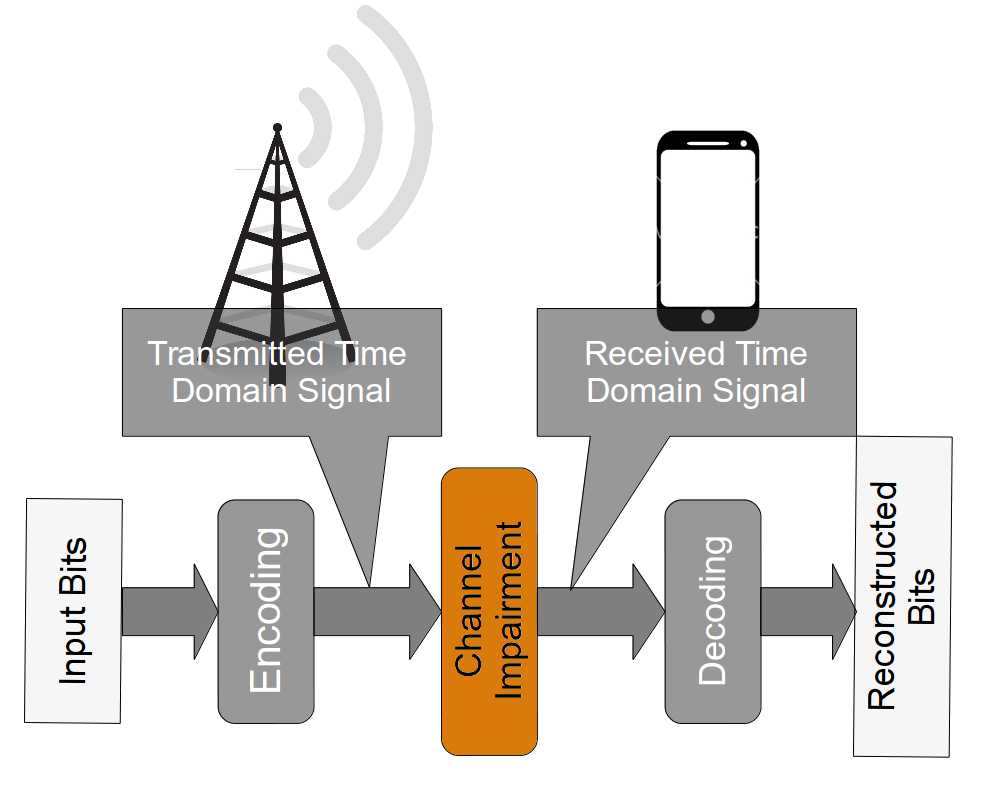}
        \caption{The basic channel auto-encoder scenario}
    \label{fig:overview}
\end{figure}

We introduce the channel auto-encoder as our primary method for learning end-to-end radio communications systems.  In its simplest form, the channel auto-encoder includes an encoder, a channel regularizer, and a decoder.  In this paper we limit our scope to binary channel auto-encoders in which the input values are binary and our output goal is to reconstruct these input bits, but the same architecture could apply to the encoding of real valued signals as well.  Figure \ref{fig:overview} provides a simple high level illustration of this model.

Auto-encoders \cite{hinton1994autoencoders} provide a powerful method for performing unsupervised learning.  They optimize reconstruction loss through a series of representations typically using mean squared error (MSE) and stochastic gradient descent (SGD) to improve regression quality.  This has been shown to work well in numerous domains, we have demonstrated the viability on radio signals in \cite{o2016unsupervised}.

Numerous enhancements have been introduced which expand on the basic autoencoder by using layers such as convolutional layers \cite{convae} to force shift invariance and reduce the number of network parameters, and the use of regularizers such as L1 and L2 regularization, input noise \cite{dnae} and dropout \cite{dropout}.  These regularizers are general purpose and well suited to numerous types of data but we also supplement them with domain specific regularizing effects.

For each measurement we use a dataset of 100,000 examples, each 128 random bits, this is a minuscule fraction of the complete input code space ($2^128$), but sufficiently allows our network to converge in most cases.  We partition this dataset into 80\% training and 20\% test/validation.

\section{Loss in Channel Auto-encoders}

Mean squared error is a commonly used loss function in auto-encoders for regression and we apply it here on discrete bits in ${0,1}$.  For hard decision regression we can use a hard sigmoid output layer and for soft regression we use a linear output layer.  Another approach might be regression of output bit likelihood $\ell \propto p(b_i=1)/p(b_i=0)$, with a binary slicer to obtain discrete values, as is commonly done in soft-decision wireless receivers.  Here we obtain the original input bits by slicing our likelihoods $\ell$ around the decision threshold $\gamma$.

\begin{equation}
\hat{b_i} = 
\begin{cases}
0 &\text{if } \ell_i < \gamma,\\
1 &\text{if } \ell_i \ge \gamma.
\end{cases}
\end{equation}

If we consider such a situation, MSE is not obviously the best choice because a large amplitude likelihood is ideal, not simply an exact predicted value.  When using likelihood we can compute hard decisions and error rate based on a set threshold $\gamma$.  This leads to the idea of a clipped error function.  With this intuition as a basis, we test four different loss functions 

\begin{align}
\mathcal{L}_{MSE} &= ( t - p )^2 \\
\mathcal{L}_{CLMSE} &= \left\{
    \begin{array}{ll}
        (t-p)^2 \times \mathbbm{1}{(p>0)}  & \mbox{if } t = 0 \\
        (t-p)^2 \times \mathbbm{1}{(p<1)} & \mbox{if } t = 1 \\
    \end{array}
    \right.   \\
\mathcal{L}_{CLMEE} &= \left\{
    \begin{array}{ll}
        e^{t-p} & \mbox{if } t = 0 \\
        e^{p-t} & \mbox{if } t = 1 \\
    \end{array}
    \right.   \\
\mathcal{L}_{CLMLE} &= \left\{
    \begin{array}{ll}
        (t-p) \times \mathbbm{1}{(p>0)}  & \mbox{if } t = 0 \\
        (p-t) \times \mathbbm{1}{(p<1)} & \mbox{if } t = 1 \\
    \end{array}
\right.
\end{align}
where $\mathbbm{I}{(\cdot)}$ is the indicator function.

\begin{figure}
    \centering
    \subfloat[Loss Functions]{{\includegraphics[width=6cm]{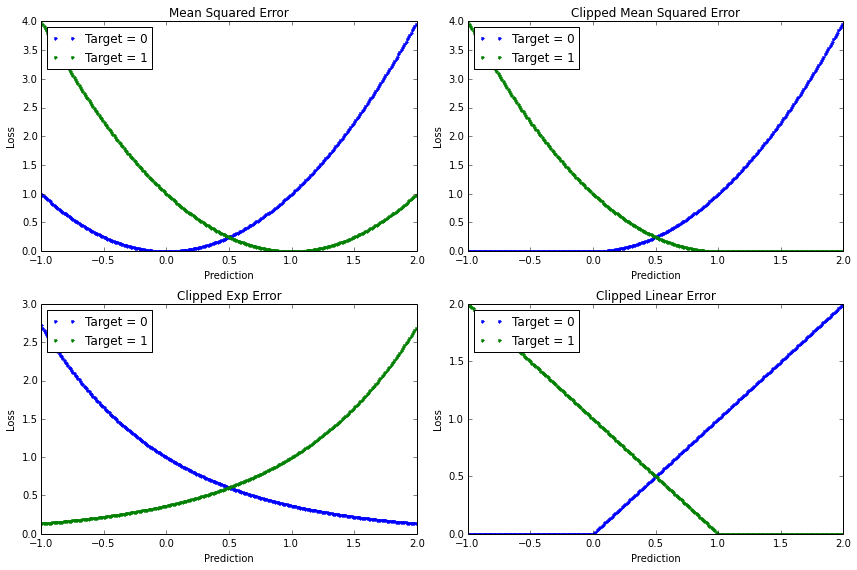} }}%
    \qquad
    \subfloat[Reconstruction Performance]{{\includegraphics[width=6cm]{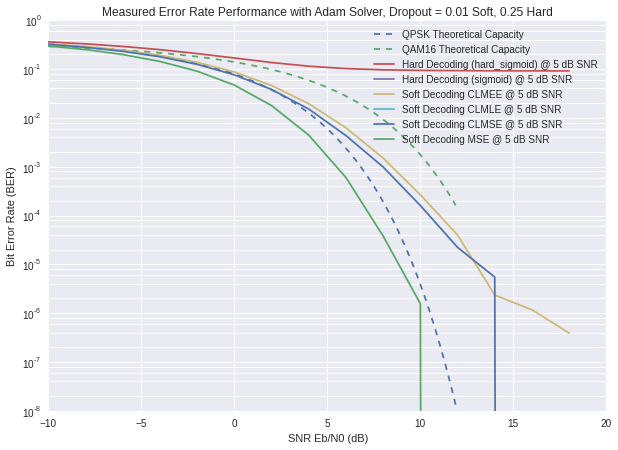} }}%
    \caption{[a] Candidate loss functions on binary input targets; and [b] AWGN Bit Error Rate Performance vs SNR of Different Loss Functions while training on Channel-AE at 5dB SNR }%
    \label{fig:loss}%
  \centering
\end{figure}

The $\mathcal{L}_{MSE}$ is the classic MSE function, which provides an equal penalty proportional to the squared distance between the predicted and true values.  In order to penalize loss only when the predicted values were on the opposite side of the threshold $\gamma$, we implement the clipped MSE loss $\mathcal{L}_{CLMSE}$.  This penalizes predicted values proportional to the square distance from the true value only for values which fall on the wrong side of the threshold $\gamma$.  The intuition is similar for the clipped linear loss function $\mathcal{L}_{CLMLE}$, except the loss is penalized linearly.  The final loss function is the exponential loss function, $\mathcal{L}_{CLMEE}$.  This penalizes the loss exponentially.  The crucial difference between this and the clipped linear and squared loss functions are that it never allows the loss to go to zero.  Our hypothesis behind this is that it would encourage the solver to encourage high likelihoods, pushing the predicted values as far away as possible from the threshold.  We compare these loss functions by measuring the bit error rate (BER) performance of the channel auto-encoder after training.  Results are shown in Figure \ref{fig:loss}.  

We evaluate BER performance with both the RMSprop \cite{rmsprop} and Adam \cite{adam} solvers at several learning and dropout rates and obtain slightly better performance with the Adam solver.  The results show that the best performance is achieved with a soft decoding receiver using the $\mathcal{L}_{MSE}$ loss.  This is somewhat counter intuitive as it needlessly penalizes higher likelihoods, but we still believe there is future value in clipped loss functions as we observed vaster training with $\mathcal{L}_{CLMSE}$ leaving open the potential for use in pre-training in perhaps larger networks.

\section{Domain Specific Regularization}

Radio signal propagation is a heavily characterized phenomenon which we can generally model well.  We select several of the most common impairments that occur during over the air transmission in any wireless system and build channel regularization layers to evaluate reconstruction learning capacity and learning difficulties under each.  Initially considered are the effects of:

\begin{itemize}
    \item $R_{noise}$: Additive Gaussian thermal noise
    \item $R_{toa}$: Unknown time and rate of arrival
    \item $R_{foa}$: Carrier frequency and phase offset
    \item $R_{h}$: Delay spread in the received signal
\end{itemize}

Gaussian noise and dropout are commonly used regularizers \cite{dropout} \cite{raviv1996bootstrapping}, however the typical use is at training time only, removing them at evaluation time.  Instead, our system must continue to cope with noise at evaluation time emulating expected channel impairment effects as we measure the network's performance.  In radio communications, additive noise is widely used to model this and referred to as Additive White Gaussian Noise (AWGN), in our implementation add a real Gaussian random variable $N$ to each in-phase (I) and quadrature (Q) sample component passed through the channel where $N \sim N(0,\frac{ 10^{(-SNR_{dB}/10.0})}{\sqrt{2.0}})$.  A normalization layer before this which normalizes the average power incoming activations to 1.

Unknown time and rate of arrival has a direct equivalence to the computer vision domain, here our regularizer applies an unknown shift and scaling in the time domain which our encoding must cope with.  In radio systems, this occurs as radio propagation times vary and clocks on distributed radio systems are typically not synchronized to some high accuracy out of band reference.  We choose a time shift of $\theta_t \sim N(0,\sigma_t)$ and a time-dilation rate $\theta_t' \sim N(1,\sigma_t')$.

Carrier frequency and phase offset does not have an equivalent to out knowledge in the vision domain, the best example would be if RGB channels were considered orthonormal basis functions and an arbitrary rotation were performed between them for each transmission.  In radio, sampling is typically done using complex baseband representation, and in this representation, unknown offsets in center frequency and absolute phase of arrival due to unsynchronized oscillators on transmitter and receiver as well as Doppler shift, result in static or linear polar mixing of the the two complex base-band components (in-phase, I, and quadrature, Q).  To simulate a real system, this layer randomly picks a phase $\theta_f \sim U(0,2\pi)$ and a frequency offset or linear phase ramp $\theta_f' \sim N(0,\sigma_f)$, where $\sigma_f$ is our expected center frequency offset error due to independent drifting oscillators.

Lastly, Delay spread in the received signal simulates the arrival of numerous delayed and phase shifted copies of a signal arriving at the receiver.  Since this is simulated as a linear system and we assume stability over a single sample time window, we can choose a random non-impulsive channel delay spread filter and convolve it with the input signal to obtain an output which has been spread in time linearly according to a random channel response.

We implement each of these regularizers as layers in Keras such that they can be easily applied to encoded signal representation of the auto-encoder. 

\section{Network Structure Selection and Evaluation}

\begin{figure}
    \centering
    \subfloat[Auto-encoder Construction]{{\includegraphics[width=6cm]{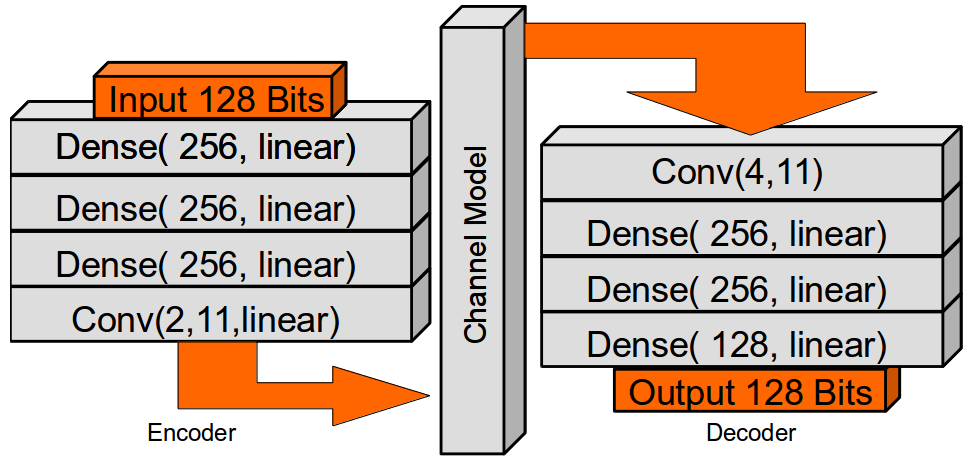} }}%
    \qquad
    \subfloat[Candidate Performance]{{\includegraphics[width=6cm]{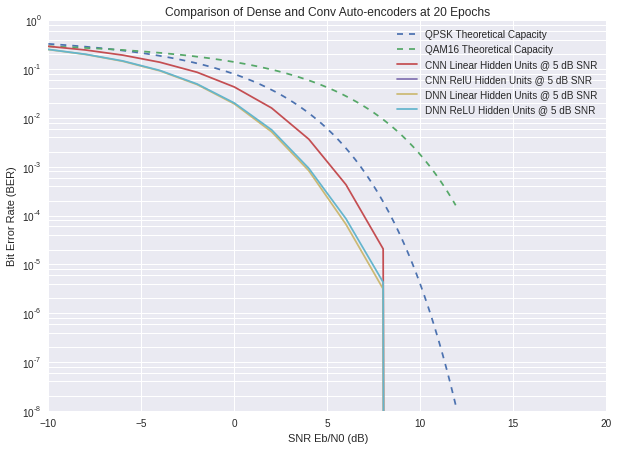} }}%
    \caption{[a] Encoder and Decoder structure used; and [b] Comparison of CNN and DNN architectures on AWGN only  }%
    \label{fig:netsel}%
  \centering
\end{figure}

We compare several network structures for our auto-encoder, including a deep dense neural network (DNN), and a convolutional neural network (CNN) with dense hidden units shown in figure \ref{fig:netsel}a.  In both cases we consider several activation functions for our hidden units, while we use a linear regression layer on the output of the encoder and a linear layer on the output of the decoder (for soft decoding), or a hard-sigmoid activation on the output of the decoder (for hard decoding).   

To evaluate the quality of each learned representation, we focus on a metric commonly used in the communications domain.  That is the bit error rate (BER) as a function of the signal-to-noise ratio (SNR), an important metric which characterizes how reliably a system can communicate bits of information as received signal power falls off.  We compare our results with the commonly used QPSK and QAM-16 expert modulation schemes which have long been used \cite{oetting1979comparison}.  The performance of each of these in terms of SNR vs BER can be expressed analytically in a purely Gaussian channel using the following expressions in equations \ref{eq:qpskber} and \ref{eq:qamber} which integrate the Gaussian error functions under the geometric conditions of a bit error.  These benchmarks are shown as dotted lines in our capacity curves to provide a comparison to widely used modern day expert modulation performance.

\begin{multicols}{2}
\begin{equation}\label{eq:qpskber}
\textbf{QPSK:} P_b = \frac{1}{2} \erfc{ \left( \sqrt{ \frac{ E_b }{ N_0 } } \right) }
\end{equation} \break
\begin{equation}\label{eq:qamber}
\textbf{QAM16:} P_b = \frac{3}{8} \erfc{ \left( \sqrt{ \frac{ 4 E_b }{ 10 N_0 } } \right) }
\end{equation}
\end{multicols}

The other important aspect of a communications scheme is that of information density, or how many bits per second can be transmitted over a fixed bandwidth channel at a specific SNR.  This is known as spectral efficiency.  For this work we maintain a constant number of hidden units per bit, and vary the SNR of these hidden units on average, keeping our spectral efficiency constant while comparing BER performance.

For AWGN we obtain best performance using a DNN with linear units, however the free parameter count in this network is significantly higher, training times are longer, and performance relies heavily on on the lack of any delay spread and the independence of each hidden unit, it does not handle channel regularizers well.  This performance comparison is shown in figure \ref{fig:netsel}b.

\subsection{Effects of Training SNR}

\begin{figure}
    \centering
    \subfloat[Performance vs Training SNR]{{\includegraphics[width=6cm]{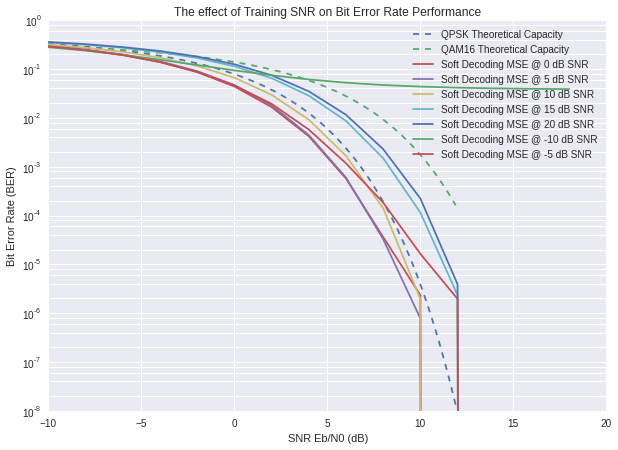} }}%
    \qquad
    \subfloat[Performance vs Training Dropout]{{\includegraphics[width=6cm]{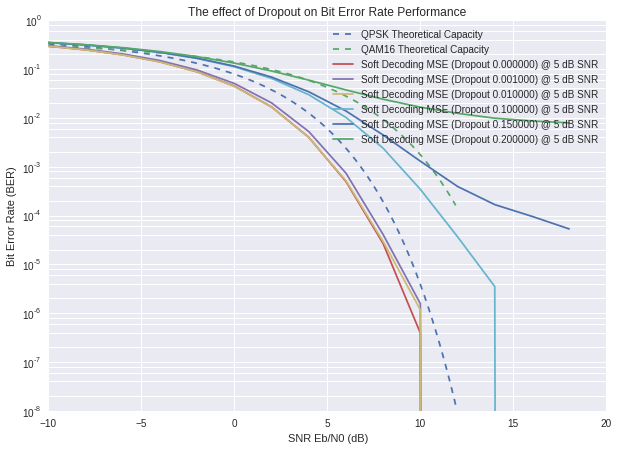}}}%
    \caption{[a] Effects of training set regularisation SNR on resulting encoding error rate performance; and [b] Effect of dropout rate on error rate  }%
    \label{fig:trainsnr}%
  \centering
\end{figure}

The SNR of the signal representation in the channel through the AWGN regularizer has a significant effect on the ultimate BER vs SNR performance of the resulting channel encoding scheme.   In figure \ref{fig:trainsnr}a we show a number of different performance plots resulting from training a CNN at a range of different training SNR.  We tend to obtain best system performance when training at low positive SNR values such as 5dB.  This is likely because here we are stressing the hidden units to represent as much information as possible to perform reconstruction, where as at negative SNR that information is too distorted to learn effective means to recover much of, and at high SNR less resilience is required in the network's ability to recover from channel error.

\subsection{Effects of Dropout}

The effect of dropout when training such a system is interesting and varies from typical systems in that our hidden units over the radio channel are constrained in time.  It also varies in that we have a lot of binary inputs which are uncorrelated independent bits and a lot of information which we must learn to represent.   We train with hidden units of twice the width of the input number of bits to allow for numerous competing models to form with dropout.  However, since we seek to preserve and reconstruct all of the bits independently at the output, any substantial amount of dropout seriously degrades the information conveyed through the network.  We find the best training performance to be with very non-zero levels of dropout such as 1e-3 or 1e-2.  Figure \ref{fig:trainsnr}b shows this effect of dropout on the ultimate channel reconstruction performance of the network after training.  It is also important to note that in each case, validation loss reaches roughly the same MSE loss value after training (within a factor of 2), but the ability of the encoding to generalize to higher SNR, lower BER encoding performance is effected (on the order of 20dB 

\subsection{Effects of Channel Delay Spread}

Delay spread varies in communication systems, from wire-line and space based wireless systems which can sometimes have very short impulsive channel responses to HF and dense multi-path wireless systems which can have long delay spreads.  We model this by introducing a regularization layer which takes the form:

\begin{equation}
    out = conv(in, h) : h \sim U(-1,+1,shape=[n_{taps}])
\end{equation}

That is, for each example through the wireless channel, we convolve a new random filter with given length: $n_{taps}$.  This is a pessimistic view for a real system since $h$ really has a more defined probabilistic envelope, but we use this simplified expression for the scope of this work.  

\begin{figure}
    \centering
    \subfloat[Impact of Delay Spread]{{\includegraphics[width=6cm]{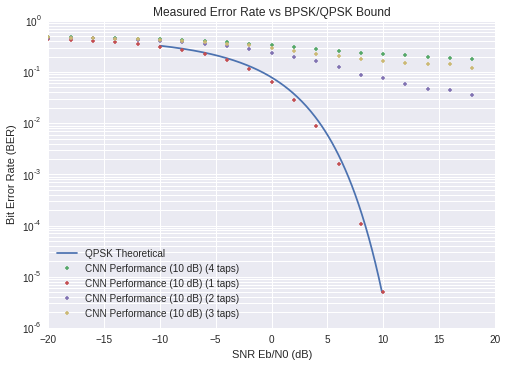} }}%
    \qquad
    \subfloat[Impact of Random Initial Phase ]{{\includegraphics[width=6cm]{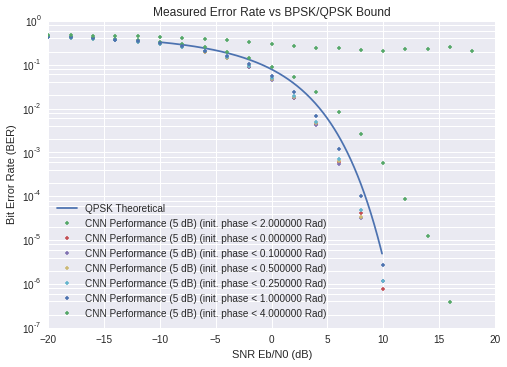}}}%
    \caption{[a] Effect of delay spread on the system without attention; and [b] The effect of random initial phase on system performance }%
    \label{fig:delayperf}%
\end{figure}

\subsection{Effect of Channel Frequency and Phase Variance}

Frequency and absolute phase of arrival of a waveform at the waveform are equivalent to a transform of the following form.  We have a transmitted input signal of shape x = [2,$n_\textbf{time samples}$], where I and Q are on separate channels.   A random phase results in the effect as follows.

\begin{equation}
    y = exp(j \theta) * x
\end{equation}

Or kept in real terms which we can easily implement in real tensor based systems such as Keras, we have

\begin{equation}
    y = [x[0,:]*cos(\theta) - x[1,:]*sin(j \theta);
         x[0,:]*sin(\theta) + x[1,:]*cos(j \theta)]
\end{equation}

The effect for frequency offset is similar except that phase becomes linearly time varying as shown below.

\begin{equation}
    y = exp(j (\theta + \theta't)) * x
\end{equation}

Unfortunately invariance to this transform is not readily learned in our experimentation.  In figure \ref{fig:delayperf}b we see that once our initial phase is distributed between 0 and 2 $\pi$, we obtain extremely poor performance.  To help our system learn the additional necessary invariance to cope with these channel effects, we introduce a model for attention.

\section{Attention Models for Receiver Synchronization}

\begin{figure}
    \centering
    \includegraphics[width=0.75\textwidth]{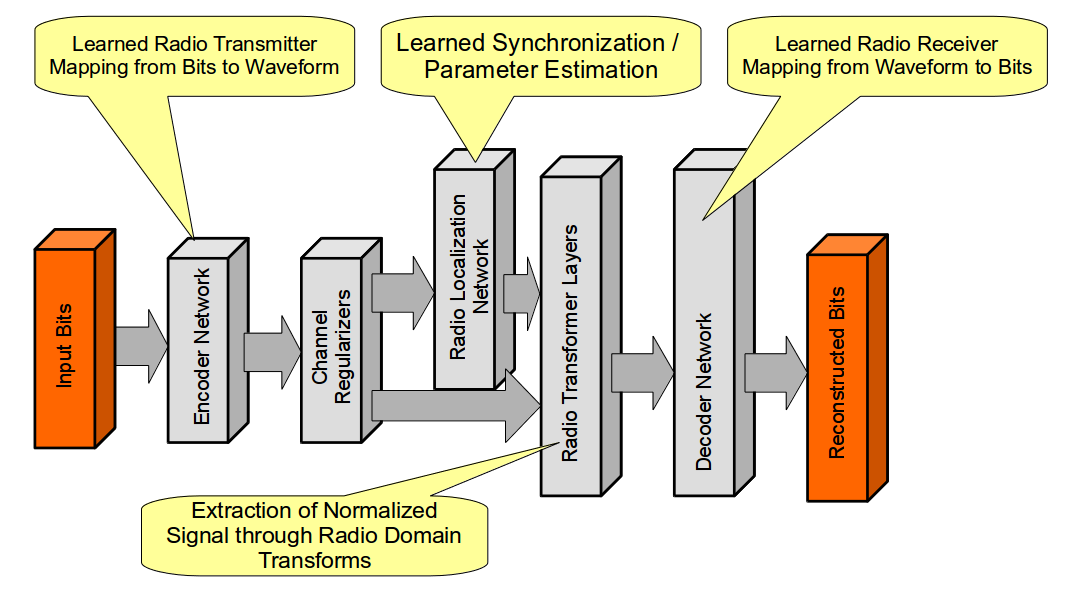}
    \caption{Our end-to-end unsupervised Radio Transformer Model. Joint learning of transmit, receive and synchronization-attention approaches to minimize bit error rate}%
    \label{fig:rtn}%
  \centering
\end{figure}

Performance decays rapidly with delay spread, as we lose the independent information within each hidden unit passing over the channel.  To address this we introduce the notion of a Radio Transformer Network for attention.  We leverage the end-to-end localization and discriminative network introduced in \cite{jaderberg2015spatial} and adapted to the radio domain in \cite{o2016radio}.  The localization network estimates parameters such as time of arrival, frequency offset, phase of arrival, and channel response taps, while transforming layers then apply it to the received data.

In the cases of phase and frequency offset, as well as time offsets, and non-impulsive channel responses, we implement a specific radio transformer algorithm as a layer in Keras for each.  Our localization network then becomes a parameter estimation and regression network for these transforms followed by a decoder.  This network then presents an entire comprehensive scheme for learning end to end encoding, decoding, and synchronization networks for naively learned communications encodings.  It can be used completely unsupervised and adapt the channel encoding scheme to any specific radio regularization layer or configuration which might be appropriate for the intended radio use environment.

\section{Visualization of Learned Modulations}

\begin{figure}[!ht]
    \centering
    \includegraphics[width=0.6\textwidth]{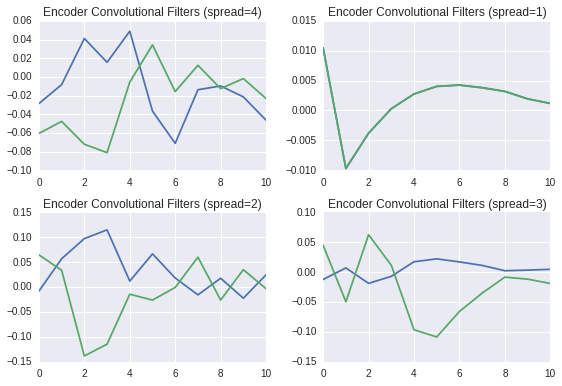}
    \caption{Encoder Convolutional Basis functions vs Delay Spread}%
    \label{fig:basisds}%
  \centering
\end{figure}

It is interesting to look at the learned convolutional features which become the representation basis for the signal over the channel.  In figure \ref{fig:basisds} we show these at 1,2,3 and 4 sample delay spread widths and they seem to take a form reminiscent of a time-frequency style wavelet basis with varying compactness in time depending on the channel delay spread.

\begin{figure}[!ht]
    \centering
    \includegraphics[width=1.0\textwidth]{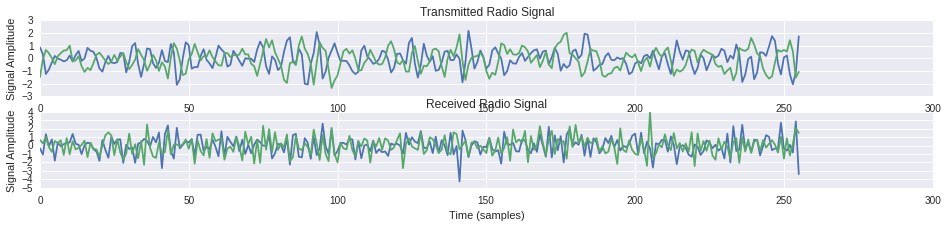}
    \caption{Example transmit and receive signal at 0dB SNR for 128 bits of information.  In-phase and Quadrature components are shown.}%
    \label{fig:txrxsig}%
  \centering
\end{figure}

In figure \ref{fig:txrxsig} we show what the same transmit and receive signal looks like throughout our system at 0dB after channel regularization.  Its interesting to note that the modulation basis here is not clearly recognizable as any existing modulation we use widely now, it seems to use at least 3 common discrete levels, but potentially encodes information in some mixture of time and frequency bins across the sample space.

\section{Computational Complexity}

Computational complexity of training deep neural networks can be quite high.  However, the feed forward execution of a trained system can be quite fast and compact since it largely takes the form of a set of sequential dense matrix operations.  In contrast much of what goes on to achieve channel capacity in modern day systems relies on sparse and iterative operations such as low-density parity codes and convolutional turbo codes which draw significant amounts of power in current day portable mobile devices.  Additionally tensor based processing is growing increasingly popular with the development of asic and GPU libraries to make execution of efficient dense tensor methods efficient, concurrent, and portable.  This potentially reduces the requirement in mobile devices to have dedicated asic space for operations such as error correction and increases the ability to move towards more general purpose compute architectures which can be shared with numerous other applications such as user level machine learning for vision and acoustic information.

\section{Realistic Deployment Considerations}

In the real world, there are a number of important considerations which differ from the simulation assumptions of the system we have discussed here.  

There are a number of ways a learned communications scheme could be used in a real world scenario.  Representations and transforms trained on closed form analytic channel models can be easily deployed in a offline training scenario.  In this case learned representations are optimized offline during training on the analytic channel model.  Derivatives of the channel transforms can be directly computed analytically during training, and given sufficiently accurate stable analytic models of the channel of interest, efficient representations for transport across the channel can be learned and used without any on-line adaptation.

However, since channels may vary wildly in the real world, depending on deployment location, conditions, or nearby effects, such a system could also hold great potential to be able to perform on-line adaptation and on-line learning of specialized representations which perform well for the given real world deployment scenario.   In this case there are two major considerations which much be addressed in the future for such as system.  First is that we do not have an exact analytic expression for the channel transform, and so we must rely on approximate gradients rather than direct derivative calculation, and second is that we must consider the communications cost, latency, and capacity of the error feedback channel among communicating and collaborating nodes in such a system while learning.  Methods for transporting important approximate gradient and error information back to transmitters will be critical to enabling such a system to maximize learning capacity and rate, while minimizing communications transport requirements for such a wireless distributed system.  We hope to further address these two key issues in future work.

\section{Conclusion}

From a wireless communications perspective, the potential impact of learning how to communicate information optimally based on the environment without expertly designed modulation schemes is enormous.  Although Shannon's seminal work defines the bounds of information communication over wireless channels, it does not explain how to reach those bounds.  Resultingly, billions of dollars have been spent in communications research to try to achieve these bounds, and unfortunately, although models for real world effects have been built, they do not encompass all scenarios and all possible observed effects, and the algorithmic complexity in expertly designed techniques required to near the bound is currently extremely high.  
We have demonstrated that this architecture \textbf{is} viable for the design and implementation of end-to-end learned communications systems with potential for performance rivaling modern day systems and pushing up against Shannon capacity bounds with vastly increased generalization and lowered complexity.

Radio transformer networks hold significant potential to learn synchronization methods for extracting canonical recovered signals from the channel, however we have not yet achieve success normalizing out all common channel effects.  Learning methods also hold great potential in accelerating learning and allowing for larger more capable end-to-end networks be using techniques such a curriculum learning which iteratively train the network under differing conditions such as gradually increasing channel effect complexity.  We believe the investigation of this application not only enables a powerful new application of unsupervised learning in the radio communication domain, but offers insight into the information capacity of similar deep neural networks used for many machine learning applications.


%
%


\printbibliography

\end{document}